# MODELING THE CHLOROPHYLL-A FROM SEA SURFACE REFLECTANCE IN WEST AFRICA BY DEEP LEARNING METHODS: A COMPARISON OF MULTIPLE ALGORITHMS

Daouda DIOUF and Djibril Seck

Laboratoire de Traitement de l'Information (LTI) – ESP –
Université Cheikh Anta Diop de Dakar
BP: 5085 Dakar-Fann (Sénégal)

**ABSTRACT**

*Deep learning provide successful applications in many fields. Recently, machines learning are involved for oceans remote sensing applications. In this study, we use and compare about eight (8) deep learning estimators for retrieval of a mainly pigment of phytoplankton. Depending on the water case and the multiple instruments simultaneously observing the earth on a variety of platforms, several algorithm are used to estimate the chlolophyll-a from marine reflectance.By using a long-term multi-sensor time-series of satellite ocean-colour data, as MODIS, SeaWifs, VIIRS, MERIS, etc..., we make a unique deep network model able to establish a relationship between sea surface reflectance and chlorophyll-a from any measurement satellite sensor over West Africa. These data fusion take into account the bias between case water and instruments.We construct several chlorophyll-a concentration prediction deep learning based models, compare them and therefore use the best for our study. Results obtained for accuracy training and test are quite good. The mean absolute error are very low and vary between 0,07 to 0,13 mg/m³.*

**KEYWORDS**

*deep learning estimators; remote sensing; chlorophyll-a*

## 1. INTRODUCTION

The sensor ocean color provides measures $\rho_{toa}$ multispectral Top Of Atmosphere (TOA) reflectance (λ) of ocean-atmosphere system in the visible and near infrared since many decades.

Reflectance $\rho_{cor}$ measured by the radiometer, corrected to Rayleigh scattering contribution, specular reflection and absorption gas is expressed as follows:
$$\rho_{cor}(\lambda) = \rho_A(\lambda) + t.\rho_w(\lambda)$$

where $\rho_A$ is the atmospheric contribution and $\rho_w$ the contribution due to the ocean, t the atmospheric transmittance.

Atmospheric correction algorithm can be used [1] to estimate $\rho_A$ and and determined chlorophyll-a from the remaining $\rho_w$.

Phytoplankton are important to marine ecosystems. Its play great role in the food web and biogeochemical cycles. Chlorophyll represents mainly the phytoplankton.

To retrieved chlorophyll-a concentration from $\rho_w$, SeaWifS sensor use OC4V4 algorithm that compute the rapports of $\rho_w$ at wavelengths 443nm, 490nm, 510nm and 555nm.

$$chl-a = 10^{(0.366-3.067R+1.930R^2+6.049R^3-1.532R^4)}$$

where $R = \log_{10}(\max(\frac{\rho_w(443)}{\rho_w(555)}, \frac{\rho_w(490)}{\rho_w(555)}, \frac{\rho_w(510)}{\rho_w(555)}))$

This mean that the maximum rapport value is taken.

The machine learning show strong computing power of classification and fitting capability to big data and multi-feature data [2]. In many fields as image recognition [3], search engines [4], stock price predictions [5], accurate results are obtained.

Machine learning found many application on earth remote sensing [6] especially for ocean data products.

Due to the non-linearity and complexity of the measurements made by ocean colour sensors, and taking advantage of the high-dimensional data reduction technique for the construction of high-dimensional predictors in input-output models of deep learning, we aim, with these last ones, to construct and optimize a chlorophyll-a concentration prediction model.

## 2. DATASET

Data are from ESA Ocean Colour Climate Change Initiative (Ocean_Colour_cci): Version 3.1 Data.

This collection contains version 3.1 datasets produced by the Ocean Colour project of the ESA Climate Change Inititative (CCI). The Ocean Colour CCI is producing long-term multi-sensor time-series of satellite ocean-colour data with a particular focus for use in climate studies.

Data products being produced include: phytoplankton chlorophyll-a concentration; remote-sensing reflectance at six wavelengths (412 nm, 443 nm, 490 nm, 510 nm, 555 nm, 670 nm)
Datasets are 5 days composite images of year 2014 and are taken in an area off the West Africa, between 6°N and 30°N and - 34°W and -8°W. This area is very important. It contains the senegalo-mauritanian upwelling zone.

Datasets of year 2014 are about 26 686 250 pixels. We random these datasets to avoid the overfitting and take the 5 % for train data and 1 % for test data which is used to test the prediction accuracy of the model.

We use also independent dataset from MODIS sensor for validation.

## 3. METHODS

We implemented ours models in Keras and sklearn libraries and we used many deep learning regressor methods to establish deep network relationship between sea surface reflectance. The aims is, first to shown the feasibility because of its multivalued of ocean color data, and second, to choose the best one and further explored it and their hyperparameters tuned. For each model, input are the six spectral sea surface reflectance and output is the chlorophyll-a.

## 3.1. Generalized Linear Models

Linear models are simple way to predict output using a linear function of input features.

The spectral sea surface reflectance represents the input features and are notes by $X = (x_1, x_2, ... x_n)$, output by $y$ and therefore $\hat{y}$ the predicted output.

**Linear Regression**: The library of sklearn allow us to parameterize a linear regression. It fits a linear model with coefficients w=(w$_1$...w$_n$) to minimize the residual sum of squares between the spectral sea surface reflectance X, and the chlorophyll-a concentration $\hat{y}$ by the linear approximation:

$$\hat{y}(w, X) = w_0 + w_1 x_1 + w_2 x_2 + ... + w_n x_n$$

Linear regression optimization is to minimize the cost function written as:

$$\sum_{i=0}^{M}(y_i - \hat{y}_i)^2 = \sum_{i=0}^{M}(y_i - \sum_{j=0}^{n} w_j \times x_{ij})^2$$

**Ridge regression**: The cost function is altered by imposing a penalty equivalent to square of the magnitude of the coefficients. The cost function to minimize is:

$$\sum_{i=0}^{M}(y_i - \hat{y}_i)^2 = \sum_{i=0}^{M}(y_i - \sum_{j=0}^{n} w_j \times x_{ij})^2 + \lambda \sum_{j=0}^{n} w_j^2$$

Ridge Regression is an optimization of Ordinary Least Squares Regression.

## 3.2. Generalized Ensemble Methods

Ensemble methods are algorithms that combine multiple algorithms into a single predictive model in order to decrease variance, decrease bias, or improve predictions. In others words, they are a sets of machine learning techniques whose decisions are combined to improve the performances of the overall system.

**Bagging regressor**: Bagging methods are aggregation methods. The bagging approach consists of trying to reduce the dependency between the estimators that are aggregated by building them on bootstrap samples. The algorithm is simple to implement: it is necessary to build *n* estimators on bootstrap samples and to aggregate them.

**Decision trees regressor**: Classification and regression trees models, or CART models, were introduced by [7]. A top down approach is applied to dataset. The complexity of the model is managed by two parameters: max_depth, which determines the max number of leaves in the trees, and the minimum number min_samples_split of dataset required to search for a dichotomy.

**Random forest regressor**: A random forest is only an aggregation trees dependent on random variables. For example, bagging trees (building trees on bootstrap samples) defines a random forest.
The Random Forest allows to improve the predictive accuracy and to control over-fitting. [8].
More than the number of trees n_estimators, the parameter to be optimized is the number of variables randomly drawn for the search for the optimal division of a node: max_features. Maximizing the max_features parameter can be achieved by minimizing the out-of-bag forecast error.

**Extra Trees regressor**: This class implements a meta-estimator that fits a number of randomized decision trees on various sub-samples of the dataset and uses averaging to improve the predictive accuracy and control over-fitting.

### 3.3. Support Vector Regressor Model

SVR algorithm use RBF as a kernel function. SVR minimizes the generalization error bound so as to achieve generalized, instead of minimizing the observed training error [9].

### 3.4. K-Neighbors regressor model

The k-nearest neighbors algorithm (k-NN) is a non-parametric. For regression, the output is based on the mean or median of the k-nearest neighbors in the feature space. The parameter to optimize to control the complexity of the model is the number of neighbor K.

## 4. RESULTS

In this section, we verify the generalization capability of the constructed models. Therefore a set of data were used for prediction experiments. Results are shown on table1. The ensemble models regressor are more quite fitting data with respectively test accuracy and mean absolute error (mae) predictions of 96,04% and 0,09 for bagging; 96,05% and 0,09 for random forest; 96,46% and 0,07 for extra tree and 93,78% and 0,13 for decision trees.

The linear model give us a test accuracy of about 76,06 %. We get a mae prediction of about 0,43. The K-neighbor model have a mae prediction of 0,09 and a test accuracy of 95,08.

The support vector and the ridge gave least good results with respectively test accuracy and mae predictions of 5,05% and 0,67; 24,54% and 0,8.

The distribution with a kernel density estimate on figure below show it clearly.

Tab.1: Test of accuracy and error prediction

| Regression | Mean absolute error (mg/m$^3$) | Accuracy (%) |
|---|---|---|
| Linear | 0.43 | 76,06 |
| Ridge | 0,8 | 24,54 |
| SVR | 0,67 | 5,05 |
| K-Neighbor | 0,09 | 95,08 |
| Extra Tree | 0,07 | 96,46 |
| Random Forest | 0,09 | 96,05 |
| Decision Trees | 0,13 | 93,78 |
| Bagging | 0,09 | 96,04 |

With regard to Table 1, we find that all ensemble models regression work well with larger prediction values.

In figure 1, we show kernel density estimation (KDE) to visualize frequency test data predicted by each method. A KDE is used to get a smooth estimation of the probability density function. This curve is estimated from the data, and the most widely used kernel is a Gaussian kernel. This is particularly useful in looking for a cluster of analyses in spectra of data.

We noted that extra tree, random forest, decision trees and bagging method are more robust.
For the following of this paper, we will work with the extra trees regression model to predict the target.

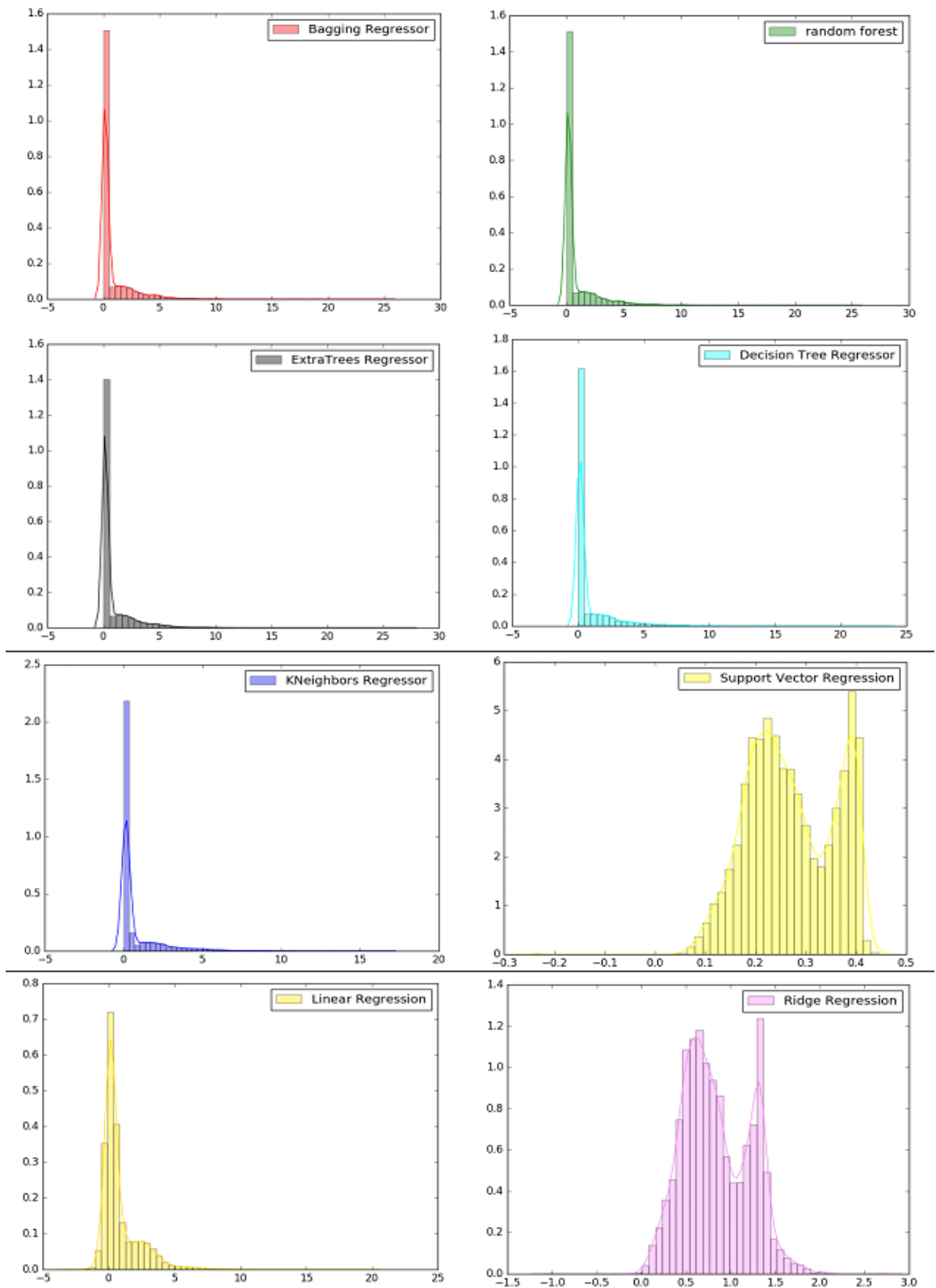

Fig. 1: Distribution with a kernel density estimation with a Gaussian kernel and a data set with 266862 sample points from a combined normal density

In figure 2 and figure 3, we plot the predicted chl-a concentration with extra trees model and the real values measured respectively for the average of period from January 16, to January 30, 2014 and for the average of period from from March 11, 2014, to March 30, 2014. For average value, we apply the algorithms to each daily image and average this daily estimate for the climatology period under study. The dataset we compare did not participate in learning phase.

The two images are very similar. Indeed, chl-a estimated for this average is equal to 95,05% with a mae of 0,09 mg/m$^3$. We see that an abundance chlorophyll-a pattern is observed near the shore and this rapidly decrease offshore. This is consistent with the literature because the upwelling area runs along the West African coast from Guinea to Mauritania. [10] and more recently [11], [12] demonstrate that the upwelling intensity is maximum in March–April.

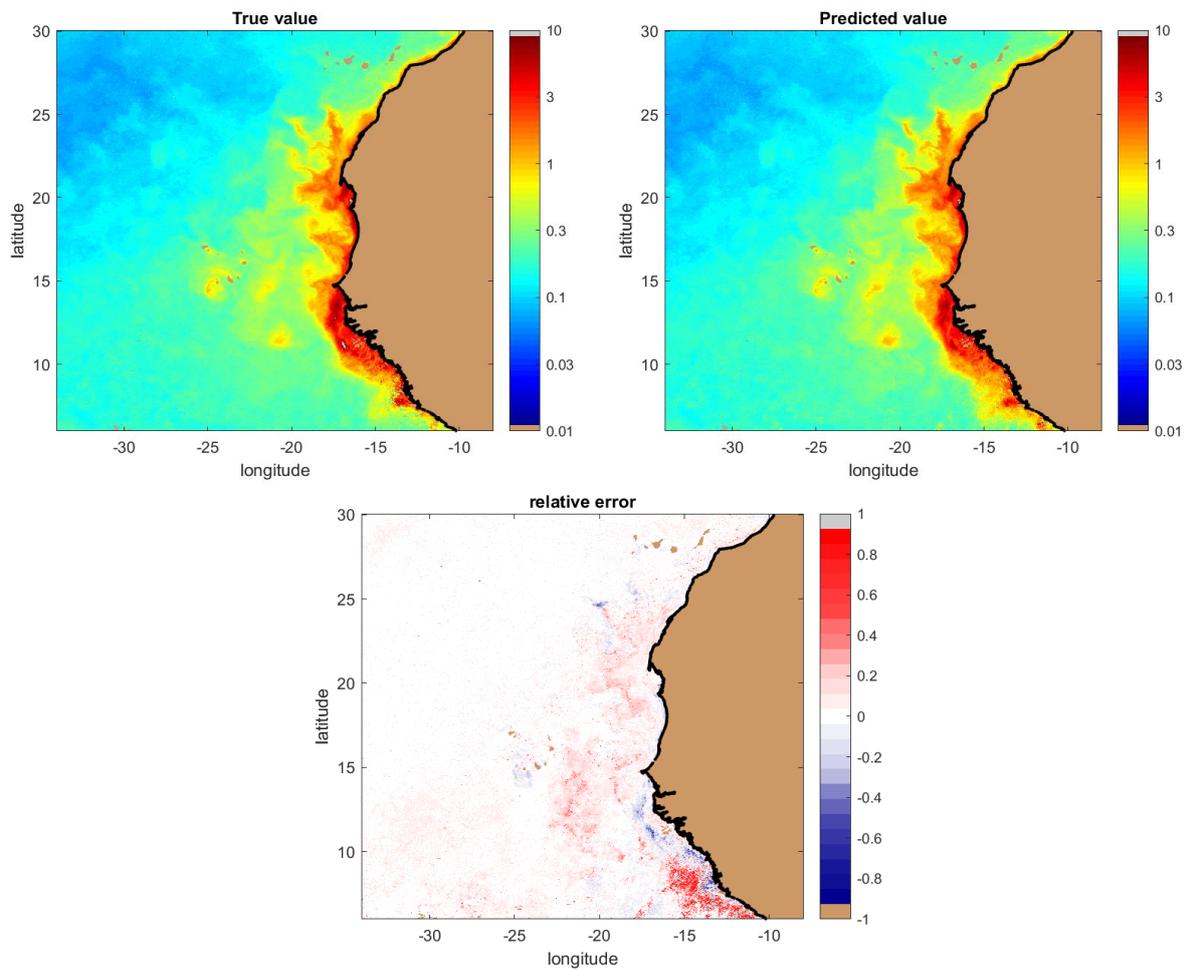

Fig. 2: The average of estimate chlorophyll-a concentration from January 16, 2014, to January 30, 2014 in mg/m$^3$ for (left) the true CCI value and (right) Extra trees prediction value. The relative error between them is represented in bottom.

International Journal of Artificial Intelligence & Applications (IJAIA) Vol.10, No.6, November 2019

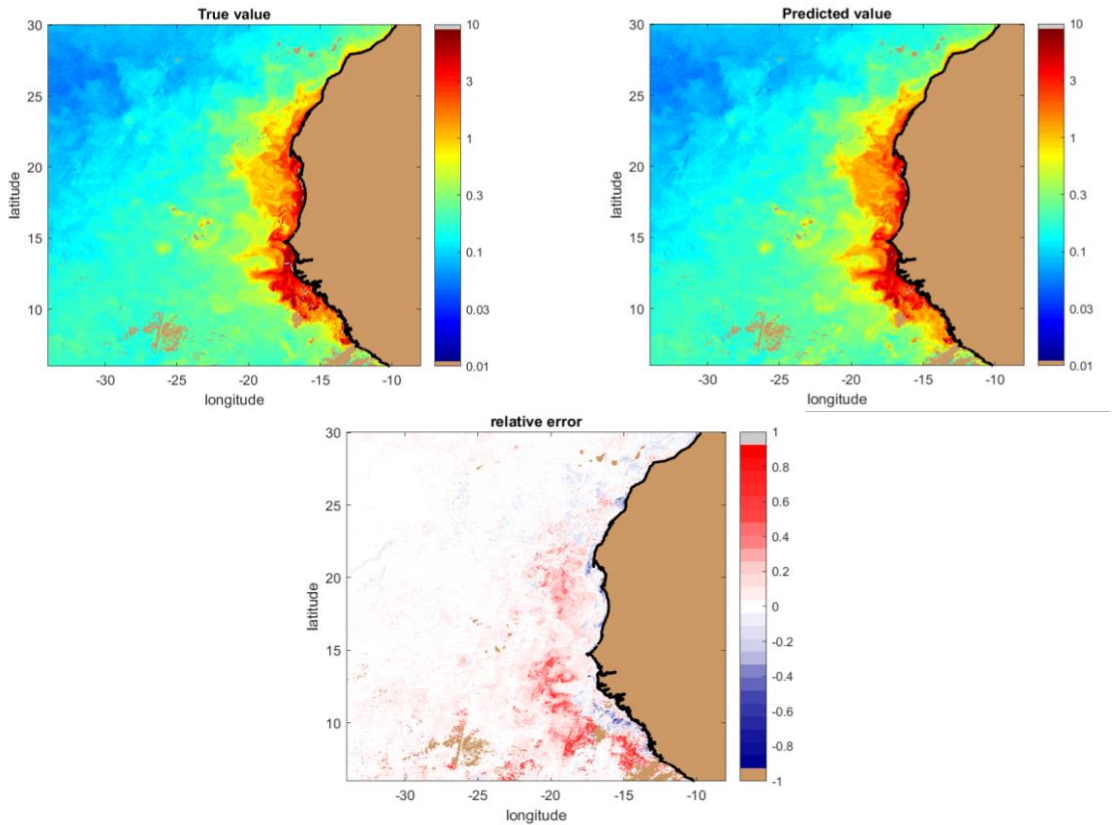

Fig. 3: The average of estimate chlorophyll-a concentration from March 11, 2014, to March 30, 2014 in mg/m$^3$ for (left) the true CCI value and (right) Extra trees prediction value. The relative error between them is represented in bottom.

By using independent dataset, comparison can be made with the chl-a predicted using from spectral reflectance of MODIS sensor with the model and the standard chl-a estimate with standard algorithm, OC3V3. This comparison shows that both methods bring out the patterns of chl-a. However the intensity of is stronger in the standard retrieval. Figure 4.

This result is very significant because it mean that according to whether we use data from several sensors, such as those used to build the different models of this work, or data from a single sensor (MODIS sensor, figure 4), the modeling capacity remains good.

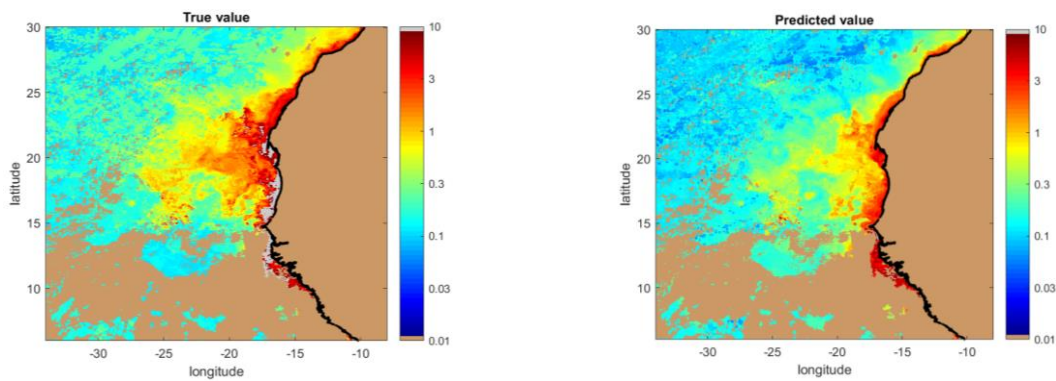

Fig.4: The average of estimate chlorophyll-a concentration from January 1, 2015, to January 16, 2015 in mg/m$^3$ for (left) the standard MODIS value and (right) Extra trees prediction value.

## 5. CONCLUSION

The high-dimensional predictors in input-output models offered by deep learning demonstrate in the work the effectiveness of chlorophyll-a retrieval from sea surface reflectance.

High accuracy are obtained on both the training and test dataset with a low mean absolute error of 0,09 mg/m$^3$ and correlation coefficient higher than 92%. The extra tree regression was the model we used. Retrievals values of chlorophyll-a are in consistence with upwelling phenomena denoted on this area. Comparison we independent value have shown satisfactory results.